\documentclass{ECOS_2021}
\usepackage{times}
\usepackage{graphicx}
\usepackage{amsmath}
\usepackage{amsfonts}
\usepackage{amssymb}
\usepackage{mathtools}
\usepackage{enumitem}
\usepackage{cite}
\usepackage{gensymb}
\makeatletter
\let\NAT@parse\undefined
\makeatother
\usepackage[hyphens]{url}
\usepackage{hyperref}
\title{\sffamily Automatic digital twin data model generation of building energy systems from piping and instrumentation diagrams}

\rmfamily

\author{Florian Stinner$^{a}$, Martin Wiecek$^{a}$, Marc Baranski$^{a}$, Alexander K{\"u}mpel$^{a}$, Dirk M{\"u}ller$^{a}$}

\address{$^{a}$ RWTH Aachen University, E.ON Energy Research Center, Institute for Energy Efficient Buildings and Indoor Climate,
Aachen,
Germany,
fstinner@eonerc.rwth-aachen.de
}

\abstract{\small Buildings directly and indirectly emit a large share of current CO2 emissions. There is a high potential for CO2 savings through modern control methods in building automation systems (BAS) like model predictive control (MPC). For a proper control, MPC needs mathematical models to predict the future behavior of the controlled system. For this purpose, digital twins of the building can be used. However, with current methods in existing buildings, a digital twin set up is usually labor-intensive. Especially connecting the different components of the technical system to an overall digital twin of the building is time-consuming. Piping and instrument diagrams (P\&ID) can provide the needed information, but it is necessary to extract the information and provide it in a standardized format to process it further.
	
In this work, we present an approach to recognize symbols and connections of P\&ID from buildings  in a completely automated way. There are various standards for graphical representation of symbols in P\&ID of building energy systems. Therefore, we use different data sources and standards to generate a holistic training data set. We apply algorithms for symbol recognition, line recognition and derivation of connections to the data sets. Furthermore, the result is exported to a format that provides semantics of building energy systems.
	
The symbol recognition, line recognition and connection recognition show good results with an average precision of 93.7\%, which can be used in further processes like control generation, (distributed) model predictive control or fault detection. Nevertheless, the approach needs further research.
}

\keywords{\small Building automation systems, Digital twin, Piping and instrument diagrams (P\&ID), Building energy system, Topology detection, Brick Schema.}

\begin{document}
\pagestyle{plain}

\section{Introduction} \label{Introduction} 

Climate change is the ultimate challenge for economics \cite{Nordhaus.2019}. Buildings have a special importance in the fight against climate change. If indirect emissions are included, buildings account for 28\% of global \cite{InternationalEnergyAgency.2019} and 36\% of European CO\textsubscript{2} emissions \cite{EuropeanComission.2018}. In Europe, the actual renovation rate of 1\% is too low to achieve the required reduction in greenhouse gases \cite{DAgostino.2017} as 3\% would be required \cite{EuropeanComission.2018}. The actual renovation rate results in a building's average lifespan between 40 and 120 years \cite{InternationalEnergyAgency.2013b}. In 2050, 75\% of the buildings in OECD countries that existed in 2010 will still be in use \cite{InternationalEnergyAgency.2013b}. Therefore, methods for minimal-invasive improvement of energy systems in existing buildings must be developed.

In particular, non-residential buildings are equipped with a comprehensive building automation system (BAS). However, their control system is incorrectly set in 90\% of total floor area \cite{Waide.2014}. About 20-30\% of total energy consumption in existing non-residential buildings with BAS can be avoided only by improved control \cite{InternationalEnergyAgency.2013b}. These novel control methods can include control generation \cite{Schumacher.2014}, (distributed) model predictive control ((D)MPC) \cite{Privara.2013} or fault detection and diagnostics (FDD) \cite{Kim.2018b} including anomaly detection \cite{Capozzoli.2018}.

Digital planning processes could remedy the deficiencies in BAS. Nevertheless, incorrectly planned controls of existing buildings are difficult to optimize. A digital twin can connect physical objects of an energy system with its virtual objects \cite{Tao.2018b} so that there is a seamless transmission of data between them \cite{ElSaddik.2018}. With help of a precise digital twin of building energy system, building automation can be optimized \cite{Fuentes.2016}. 

Machine learning applications can support to generate a digital twin. In order to use automatic control systems like energy-cyber-physical systems \cite{Schmidt.2017}, system knowledge of building energy systems (BES) and their technical building equipment (TBE) must be acquired. If this knowledge is not digitally processable in existing buildings, it must be gained in a personnel-intensive way.

Piping and instrumentation diagrams (P\&ID) provide information on how the parts of an energy system are linked, such as connecting a boiler to a distribution system. They are often not available in a computer-interpretable form. P\&ID can be digitized as cost-effective alternative for labor-intensive processing of P\&ID to increase the understanding of the system \cite{Arroyo.2016}. They are often available as a file in PDF format or as a printed or drawn plan. These plans provide valuable information for the optimization of building operation and the creation of a digital twin. The costs of optimizing existing BES can be reduced by automated extraction of information contained in the plans.

In order to control the increasingly complex control systems in a stable and optimized way, they must be permanently improved by new systems, solutions and products \cite{Isaksson.2018}. Here, a digital twin can assist operation \cite{Lydon.2019}. A format that includes domain knowledge in building services engineering, especially for analysis of BAS, is crucial to ensure that this format can continue to be used during its lifespan.

For analysis of operation of a building, the Brick Schema was developed \cite{Balaji.2018}. It contains most of necessary connections of energy systems and description of their sensors and actuators and is expected to become part of BACnet, which is one of the most widely used protocol for BAS \cite{Haynes.2018}.

The sensors, actuators and controller of a building generate data streams. These data streams also have a data stream identifier. It is usually constructed by a label structure previously defined by the building owner or building automation company. Since the label structures differ greatly in practice \cite{Gao.2018,Stinner.2018}, they must be automatically processable. For this purpose the BUDO Schema was developed as a universal data stream identifier \cite{Stinner.2018}. It can be used for existing and new buildings.

In our approach, the P\&ID is transferred to Brick Schema for further processing and to the BUDO Schema for conversion to a label structure. Building energy performance simulation (BEPS) models can then be generated from BUDO Schema and Brick Schema \cite{Stinner.2019c}. The reduction of personnel expenses in BEPS generation is crucial to ensure that BEPS can be used on a scalable basis \cite{Egan.2018} and in model predictive control \cite{Privara.2013}. The labels according to BUDO Schema are integrable into the BAS. BUDO Schema offers a possibility to name and identify data streams directly from BEPS.

In this paper, we answer whether piping and instrumentation diagrams are a valid source for the extraction of topology of existing building energy systems of non-residential buildings and their technical building equipment for modern applications in BAS such as model predictive control, building energy performance simulation (BEPS) or fault detection and diagnostics.

Our contributions are as follows:

\begin{itemize}
	\item Creation of a data set for detection of topology of various standards in plans of building energy systems
	\item Algorithm for export of P\&ID of building energy systems into a digital twin model for 
	\begin{itemize}
		\item model predictive control
		\item fault detection and diagnostics
		\item automated building energy performance simulation model generation
		\item integration into building automation system
	\end{itemize}
	\item Time tracking of all needed process steps
\end{itemize}

First, we describe the used data set. The training data set is generated from three different sources. The test data set consists of cutouts from five different P\&ID from different vendors and used standards. Based on this, we present the methodology used for symbol detection, line detection and cross detection. Afterwards, we present the results generated using our method and our data set. Finally, we discuss the limitations of our approach, but also compare them with approaches from industrial applications. We give an outlook for which applications our approach is suitable.



\section{Related work} \label{sec:related_works} 

There are few commercial solutions for digitization of inventory plans. These are mainly limited to visualization, but do not include the data transfer for automation systems \cite{Arroyo.2015}. For the history of image recognition and processing of P\&ID, we refer to the work of Moreno-Garcia et al. \cite{MorenoGarcia.2018}.

There are only few existing automatic approaches for analyzing P\&ID of BES. These used computer-aided design (CAD) or Building Information Models (BIM) models as input. Zhang \cite{YuxiZhang.2019} developed a symbol recognition for BES P\&ID based on Faster R-CNN \cite{Ren.642015}. However, this approach lacks the recognition of connections between identified symbols and their export.

Building Information Model (BIM) is a digital model and process that is used to coordinate the various trades, for example building automation. Such a digital model could be a good data source of the building energy systems of existing buildings. However, its dissemination is limited \cite{Charef.2019}. For example, a survey conducted by the German Chamber of Architects in 2017 indicated that only 12\% of surveyed architects use BIM at all, and only 47\% of these use BIM in all projects \cite{Rei&Rommerich.2017}. Thus, it can be assumed that only a very small part of the existing buildings were modeled with BIM though with CAD. Additionally, the CAD data of buildings are often not transferred to the building operator. Therefore, alternative ways for information extraction have to be found.

In buildings, various standards are used for the creation of P\&ID as BAS partly controls energy supply of industrial plants \cite{DeutschesInstitutfurNormung.200112,DeutschesInstitutfurNormung.200401,DeutschesInstitutfurNormung.1991,DeutschesInstitutfurNormung.1997,DeutschesInstitutfurNormung.1997b,DeutschesInstitutfurNormung.1997c,DeutschesInstitutfurNormung.042015,DeutschesInstitutfurNormung.042013,DeutschesInstitutfurNormung.042011,DeutschesInstitutfurNormung.012018,DeutschesInstitutfurNormung.072014,DeutschesInstitutfurNormung.042015b,DeutschesInstitutfurNormung.102015}. The symbols and designations changed over time and are not used consistently in practice. This is partly due to the harmonization of standards. This makes a universal approach to the digitization of existing P\&ID of BES into a computer-interpretable form difficult.

The approaches for the automatic processing of P\&ID shown so far concentrate on the automatic analysis of industrial plants \cite{Gellaboina.2009,Arroyo.2016,Gutermuth.2017,MorenoGarcia.2017,Martinez.2018,Tan.82020188242018,Koltun.2018b,Kang.2019,Rahul.2019,Yu.2019,Nurminen.2020,Rica.2020}. Partly, the same technical systems are considered in the cross-sectional technology as in buildings. However, these systems are often connected differently in industry than in buildings. The budget for the analysis of systems also differs significantly here. As far as the authors are aware, there is no common approach for exporting P\&ID from BES into a computer-interpretable form. 

There are different approaches for the detection of connections of symbols in technical systems. Gellaboina et al., Zhang and Nurminen et al. \cite{Gellaboina.2009,YuxiZhang.2019,Nurminen.2020} only recognize symbols of P\&ID. We have no access to the algorithms used by Arroyo et al., Bigvand and Fay, and Kang et al. \cite{Arroyo.2016,Bigvand.2017,Kang.2019} or they remain unclear.  Guthermuth and Hoernicke \cite{Gutermuth.2017} additionally use real data stream data to support topology recognition. Hoernicke et al. \cite{Hoernicke.2015} use human-machine interface (HMI) of technical systems. However, details of used algorithms remain unclear.

A geometric matching is used by Koltun et al. \cite{Koltun.2018b} for symbol detection and thus postulate that the symbols for an equipment type in different plans are very similar and they use the square scan algorithm \cite{ElHarby.2005} for line detection. However, no results exist and only one plan was involved. Moreno-Garcia \cite{MorenoGarcia.2017} apply an heuristic algorithm to improve the object and connection recognition of complex P\&ID while Yu et al. \cite{Yu.2019} use an independent process for connection detection based on pixel counting. Guthermuth and Hoernicke \cite{Gutermuth.2017} used rule-based connection detection and thus an approach that is dependent on the creation of plans and standards it contains.

Probabilistic Hough transformation \cite{Kiryati.1991} is used by Rahul et al. \cite{Rahul.2019} for connection recognition. They use fully convolutional networks (FCN) \cite{Long.2015} with neural networks for symbol recognition. However, it is not exported into a reusable format. Additionally, they achieved only 65.2\% accuracy in detecting connections in the plan.

None of the examples shown in previously mentioned approaches has transferred a P\&ID of a BES into a digital twin model using object recognition and topology extraction.

\section{Data set} \label{sec:data_set} 

Since the standards for objects in P\&ID vary in BES, we use different data sources for raw data to generate a comprehensive data set for training and testing. Raw data is defined in this work as graphical documents that contain information needed for the detection tasks but can not directly be used for creating the data set of TBE symbols or connection derivation. Instead, preprocessing is required. We use in this work:

\begin{itemize}
	\item Google Crawler image web search data \cite{Vasa.2019}
	\item data from standards that standardize TBE symbols
	\item data from P\&ID
\end{itemize}

All accessible examples of P\&ID contain textual descriptions of the individual sensors and actuators, e.g. according to DIN EN 81346 \cite{DeutschesInstitutfurNormung.2009}. This extensive description of the sensors is unlocatable in any of our considered examples of BES. In BES, P\&ID textual annotations are often based on natural language or non-standard abbreviations and are individually integrated into the P\&ID of BES. For these reasons, the text entries in the plans have not been considered in the prepared data set and used algorithm.

\begin{figure}[htbp]
	\centering
	\includegraphics[width=0.8\linewidth]{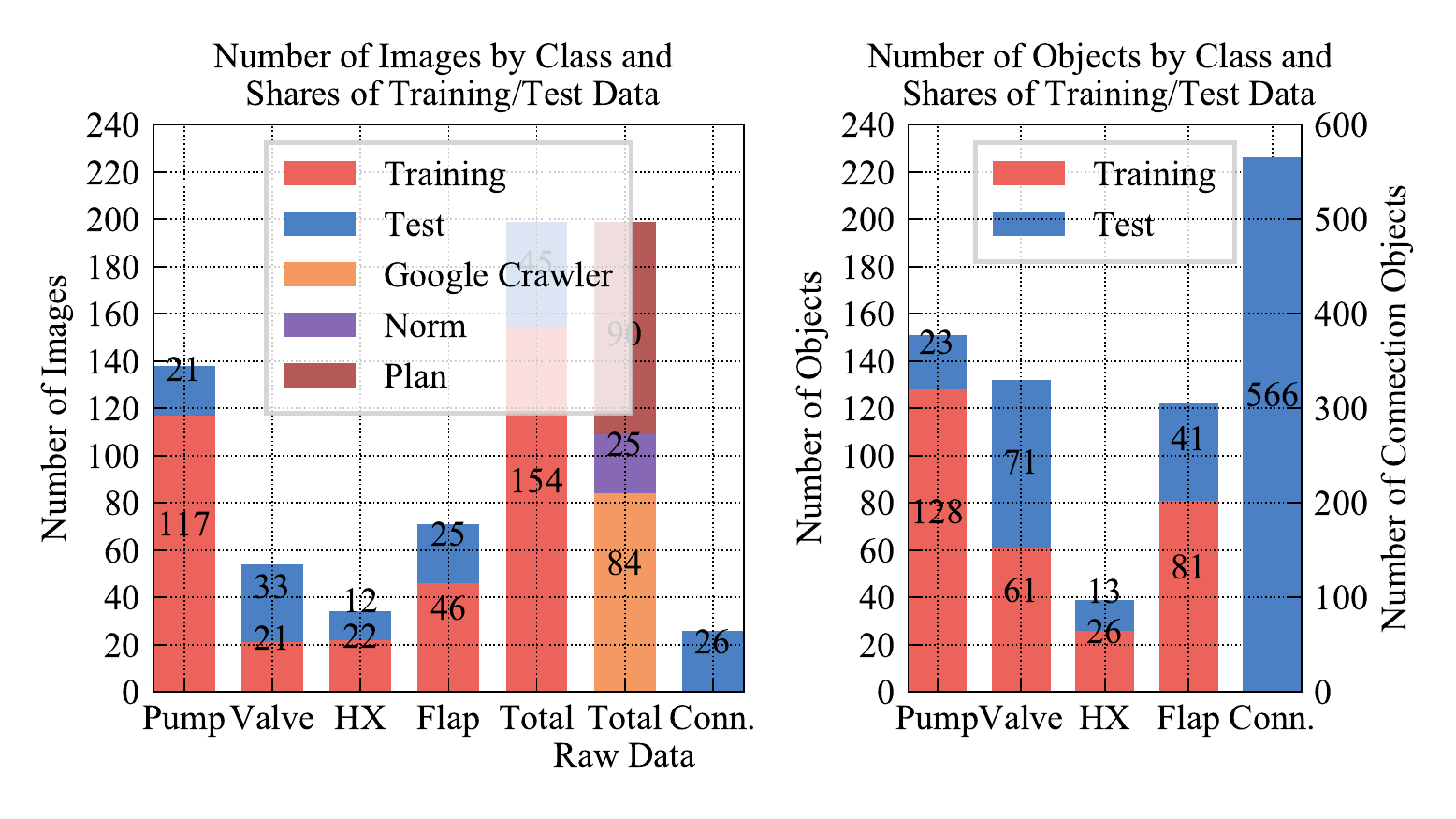}
	\caption{Histogram of the training and test data set showing number of images and objects by class (pump, valve, HX=heat exchanger, flap, total, raw data sorted by source, connection detection}
	\label{fig:HistogramImgObjbyClass}
\end{figure}

The data sets we use for object detection four TBE symbols, which are pump, valve, heat exchanger and flap. They can be found most frequently in P\&ID of BES and have a high influence on the control.
Figure \ref{fig:HistogramImgObjbyClass} shows a summary of the training and test data set.

The sum of images for all TBE symbols is not equal to the total number of images, as one image can contain different TBE symbols. The training data set contains 154 images in total, where as 84 images were extracted using Google Crawler \cite{Vasa.2019} with the following keywords:
\begin{itemize}
	\item \{"pump", "valve", "heat exchanger", "flap"\} each combined with:
	\item \{"symbol", "symbol iso", "standard symbol", "19227", "81346", "60417"\}
\end{itemize}

13 standards are used \cite{DeutschesInstitutfurNormung.200112,DeutschesInstitutfurNormung.200401,DeutschesInstitutfurNormung.1991,DeutschesInstitutfurNormung.1997,DeutschesInstitutfurNormung.1997b,DeutschesInstitutfurNormung.1997c,DeutschesInstitutfurNormung.042015,DeutschesInstitutfurNormung.042013,DeutschesInstitutfurNormung.042011,DeutschesInstitutfurNormung.012018,DeutschesInstitutfurNormung.072014,DeutschesInstitutfurNormung.042015b,DeutschesInstitutfurNormung.102015} and 25 cutouts generated from it. Further, 45 images are from plan cutouts of five P\&ID from different buildings and design companies. These P\&ID include complex energy systems such as district energy systems.

As data source for the test data we use plan cutouts. The test image data set consists of 45 images from different technical plans. The test to training data ratio is approximately 1:3. As the test data is generated with the focus on including images that can be used for the connection detection, more valve objects exist in the test than training data.

Unfortunately, there is no publicly available data set for P\&ID and for BES in particular. For creating the data set that was used for our symbol recognition training process, the following preprocessing steps were done:

\begin{itemize}
	\item Google Crawler data sorting
	\item P\&ID cutout generation
	\item norm cutout generation
	\item training data labeling
\end{itemize}

The time duration for these steps can be found in figure \ref{fig:TemporalEffortsOD1_2}.

The requirements for a test data set for evaluating the connection detection differ from the object detection. It is necessary that connections between objects are shown. Therefore, for the connection test data set only the subset that comply with this requirement was chosen. Therefore, the connection data set consists of 26 images and 566 classification elements.

\section{Method} \label{sec:method}

In Bode et al.\cite{Bode.2019b}, a toolchain is presented how raw data of building automation systems and digitization of P\&ID support new control approaches like control generation, (distributed) model predictive control or fault detection in building automation systems. In this paper, we present the digitization of P\&ID in BES in a computer-interpretable form.

In our algorithm, first, the objects within P\&ID are recognized with object detection on the basis of their symbols. Subsequently, the lines of P\&ID and crossings of these are identified. From all this information, the next step is to determine whether a connection exists between the objects. At the end, the connections are exported into a topology model for digital twin generation.

\subsection{Implementation of plan topology detection algorithms}
\begin{figure}[h]
	\centering
	\includegraphics[width=0.6\linewidth]{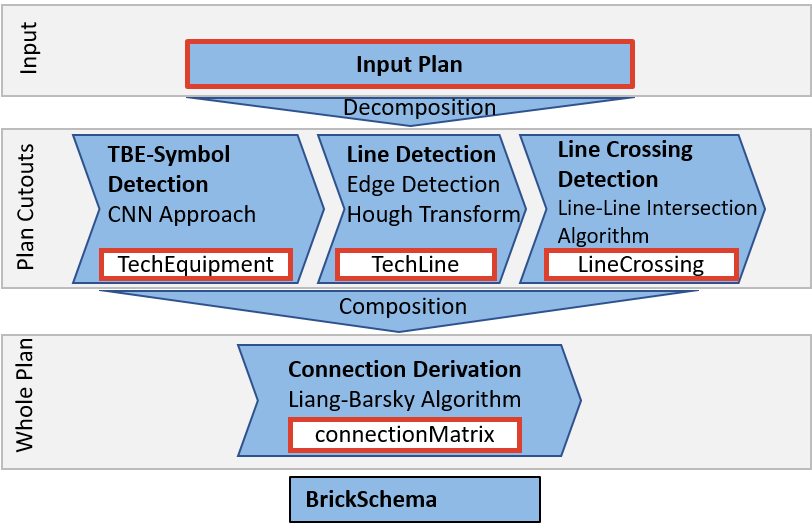}
	\caption{Schema showing the step-wise detection procedure (decomposition, symbol detection, line detection, line crossing detection, connection derivation and export function)}
	\label{fig:ProcedureSchemata}
\end{figure}

For the detection of the plan elements and topology a step-wise approach was used. The overall structure of the approach can be seen in Figure \ref{fig:ProcedureSchemata}. 

A Python-based object-oriented structure was established in our algorithm, which holds the information gathered from our algorithms. The input plan image is saved. Further, for all three core objects of the plan (TBE, technical lines and line crossings) three classes TechEquipment, TechLine and lineCrossing are established. Furthermore, a two-dimensional connectionMatrix describes the connections between all TBE.

Faster R-CNN \cite{Ren.642015} is used for symbol recognition. However, Faster R-CNN  has problems to detect smaller objects \cite{Eggert.2017}, thus why it is unable to directly recognize all the symbols on P\&ID. So we decompose automatically the P\&ID into individual cutouts that are recognized by Faster R-CNN and then reassembled. The TBE symbols, the polylines and locations of line crossings are extracted. The plan cutout information are composed to the whole plan level. A connection derivation algorithm determines TBE connections. Last, the information about TBE objects and connections between them are saved in a data interface.

\subsubsection{Decomposition and composition of technical plan}
Extracting whole plans from a PDF file as image file would result in high image sizes as a minimum dimension for the symbols must be fulfilled. Our object detection classifier could not handle this high image sizes and the high number of symbols to be detected. Therefore, we developed an approach to decompose the whole plan into cutouts. The cutout allocation is saved, so that after extracting the core elements from the cutouts, the information can be composed again and form the whole plan.


\subsubsection{Object detection convolutional neural network approach}
\label{sec:CNNmethod}
Since the symbols in BES are similar, but not exactly the same, a template matching approach is not appropriate where an exact symbol is sought. A first test with template matching could confirm this assumption (F1 score $<$ 70\%). Symbols can also be rotated or skewed. Therefore, a general approach is needed for BES, which also does not recognize exact copies of its training data symbols.  Faster R-CNN \cite{Ren.642015} provides this approach. Therefore, we developed a function based on it to detect the TBE symbols.

A function was developed for detecting the technical equipment symbols using Faster R-CNN \cite{Ren.642015}. To use the function, first a CNN based object classifier was generated by training the R-CNN using the training data set previously described. The labeled training images contain the position and class information for all objects. The training was done using a TensorFlow script \cite{Abadi.2015}. Around every five minutes the training routine periodically saves checkpoints which contain the current state of the trained classifier. The total time duration for the training can be found in figure \ref{fig:TemporalEffortsOD1_2}. After training is done, checkpoint with the highest number of steps is used to generate the inference graph. The inference graph contains the trained object detection classifier.

\subsubsection{Line detection algorithm}
\label{sec:LineDetection}
All technical equipment are connected via polylines in P\&ID in BES. Detecting the lines is therefore vital to derive the connection information. First, the image is converted to a binary image, which is used to identify edges inside the image. Next, the lines are identified using the Hough-Transform \cite{Hough.1962}.

\subsubsection{Line crossing algorithm}
\label{sec:FindLineCrossing}
\begin{figure}[h]
	\centering
	\includegraphics[width=0.65\linewidth]{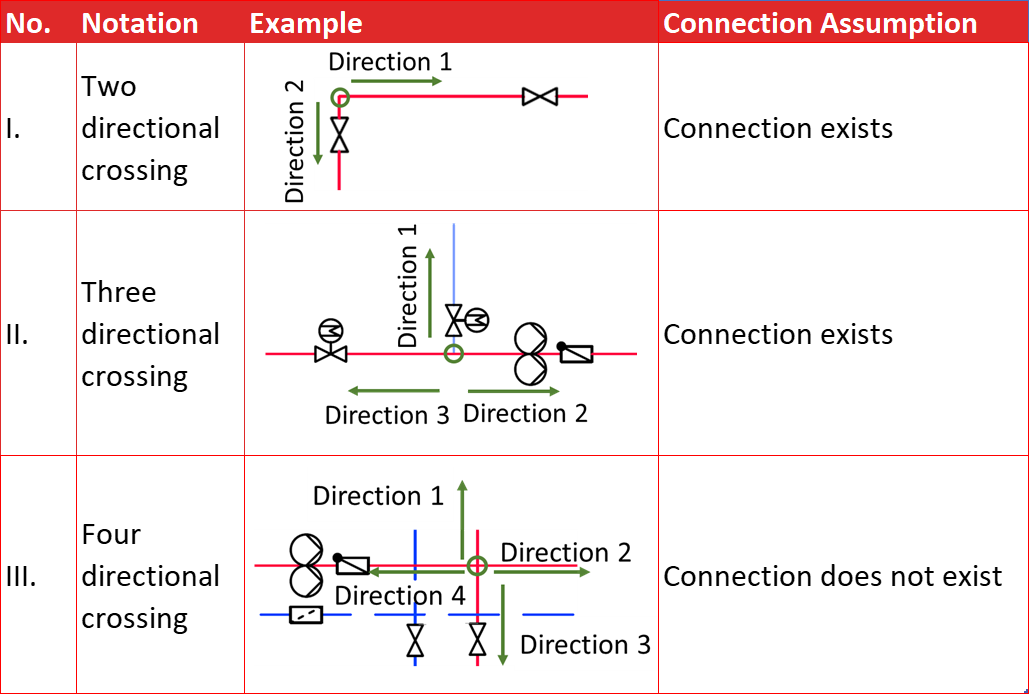}
	\caption{Typical line crossing types in piping and instrumental diagrams in building energy systems (two directional crossing, three directional crossing and four directional crossing)}
	\label{fig:LineCrossingTypes}
\end{figure}

Line Crossings are locations inside the plan image, where straight lines intersect with other straight lines. They represent a change of direction of a connection line. For the plans we used, we found that a connection exists if two or three lines leave the line crossing and no connection exists if four lines leave the line crossing. Figure \ref{fig:LineCrossingTypes} shows examples for two, three and four directional crossings and the derived connection assumption.

Every detected line is compared to all other lines for intersection identification. Consequently, the algorithm mainly does one task, which is finding the intersection of two lines. We used the approach described in \cite{Weisstein.2019} where a line-line intersection can be found using determinants.

\subsubsection{Connection derivation algorithm}
\label{sec:TopologyDetection}
\begin{figure}[h]
	\centering
	\includegraphics[trim = 20mm 8mm 16mm 22mm, clip, width=0.55\linewidth]{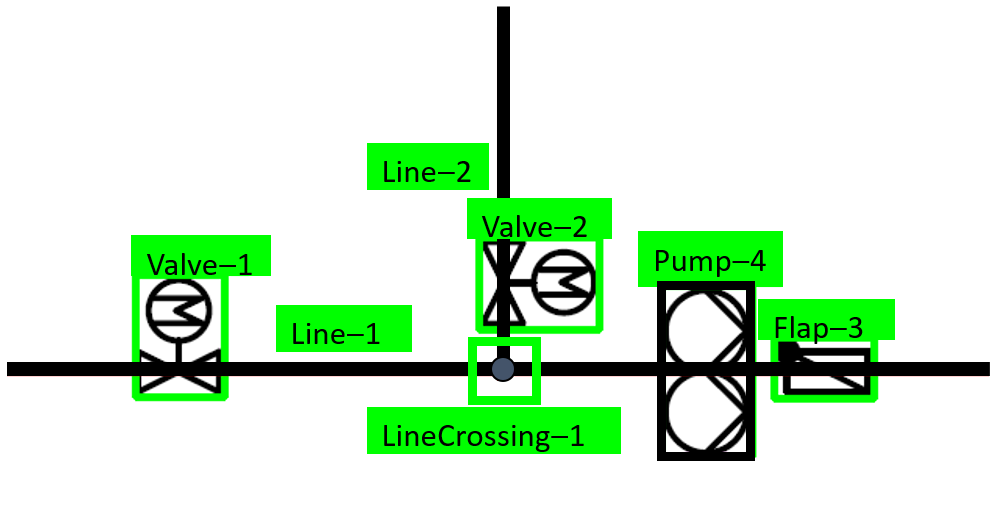}
	\caption{Connection derivation procedure with detected symbols (Valve-1, Valve-2, Pump-4, Flap-3), detected lines (Line-1, Line-2) and line crossings (LineCrossing-1)}
	\label{fig:ConnectDerivation}
\end{figure}

As basic information the connection derivation algorithm uses the core objects described in previous sections and join them in order to derive the topological connections. The procedure can be seen in Figure \ref{fig:ConnectDerivation}, exemplarily shown for finding the connecting TBE symbols for "Pump-4".
First, the lines which cross that symbol are identified. This is done using the Liang-Barsky-Algorithm \cite{Barsky.1992}. Then, the neighboring elements for this symbol are detected. This is done by searching for the closest elements that is on the same line as the symbol. In Figure \ref{fig:ConnectDerivation} this is "LineCrossing-1" for the left neighbor. Therefore, the TBE neighbors depend on the directional information of the line crossing, according to Figure \ref{fig:LineCrossingTypes}. As this is a three-directional crossing, all symbols linked via this crossing count as connection. So the neighboring TBE symbols for "Pump-4" can be determined as "Valve-1" and "Valve-2". The procedure will be done for all TBE symbols on the test image.

\subsection{Data interface}
\label{subsec:DataInterface}

Industrial Foundation Classes in Version 4 (IFC4) is the actual open source standard for BIM. It has only a limited descriptive capability for sensors and actuators and their relations \cite{Lange.2018}. Their descriptiveness is used even less frequently in real-life projects \cite{Lange.2018}.

None of the standards used for models by \cite{Koltun.562019592019} and \cite{Arroyo.2016} includes specialized domain knowledge from BAS and BES.



For these reasons, we use the domain specific ontology-based Brick Schema \cite{Balaji.2018} and label-based BUDO Schema \cite{Stinner.2018} to create a digital twin of BES systems as BES currently lack pervasive standards for automation \cite{Krutwig.2019}.

\subsection{Used computer hardware and software}
For all Computer Vision tasks a HP Pavilion x360 convertible laptop, with 8 GB RAM, Intel Core i5-8250U CPU with 1.8 GHz and Nvidia GeForce MX130 graphic processing unit (GPU) was used. The computation capability for this GPU is 5.0, which makes it possible to use the Nvidia Compute Unified  Device Architecture (CUDA) toolkit, that enables implementing parallel computing applications. For this work CUDA toolkit 10.0 was used for the training of the CNN. All programming tasks in this approach were done with the programming language Python.

\section{Results} \label{sec:results}

To validate the proposed detection algorithms, binary classification is used. Important performance assessment parameters derived from binary classification are recall, precision, F1 score, accuracy and average precision (AP) \cite{Nanopoulos.2001}.

\subsection{Object detection performance}
\begin{figure}[h!]
	\centering
	\includegraphics[width=0.7\linewidth]{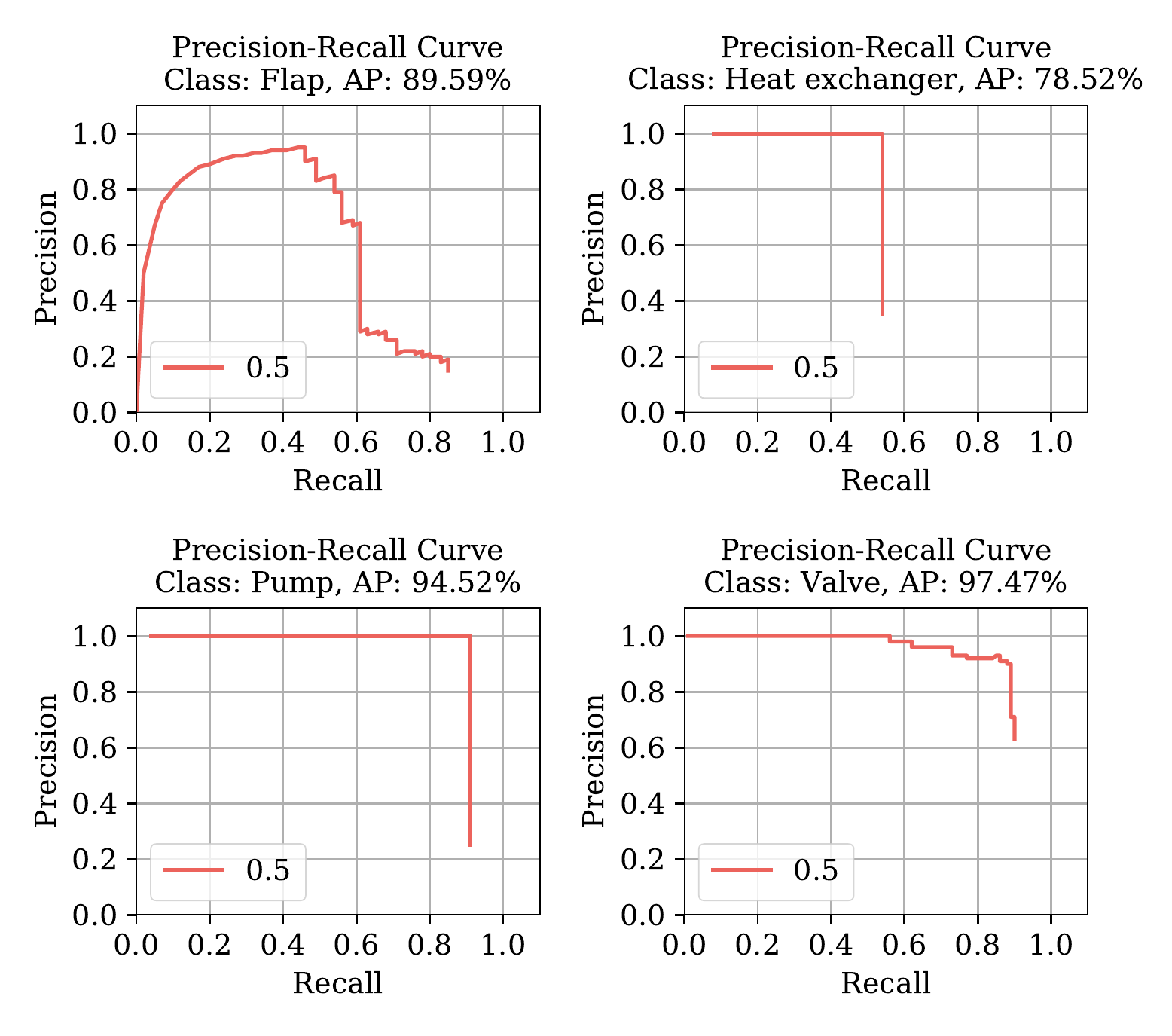}
	\caption{Precision-recall-curve for classes flap, heat exchanger, pump and valve, using IoU-threshold 0.5}
	\label{fig:all05IoU_TF}
\end{figure}

Figure \ref{fig:all05IoU_TF} shows the precision-recall curves for each of the four object classes. The previously defined object detection test data set is used. The target state for the detection is the ground truth which is defined for every test image. The curves are generated using an Intersection over Union (IoU)-threshold \cite{Rezatofighi.2019} between the ground truth and the detection bounding boxes of 0.5, which led to the highest precision and recall values.

All curves show the typical course of the graph, high precision values for low recall values and a decreasing precision for higher recall. For all curves, the precision stays above 90\% until a recall of around 80\%. For all classes, this leads to AP values of around 80\% or higher. The highest AP is reached for the class valve with a value of 97.5\%. Even at a recall of 1, the precision is at around 90\%. Consequently, all valve symbols labeled as ground truths were detected, while only 10\% of the predicted detections were false positives.

\begin{figure}[ht]
	\centering
	\includegraphics[width=0.8\linewidth]{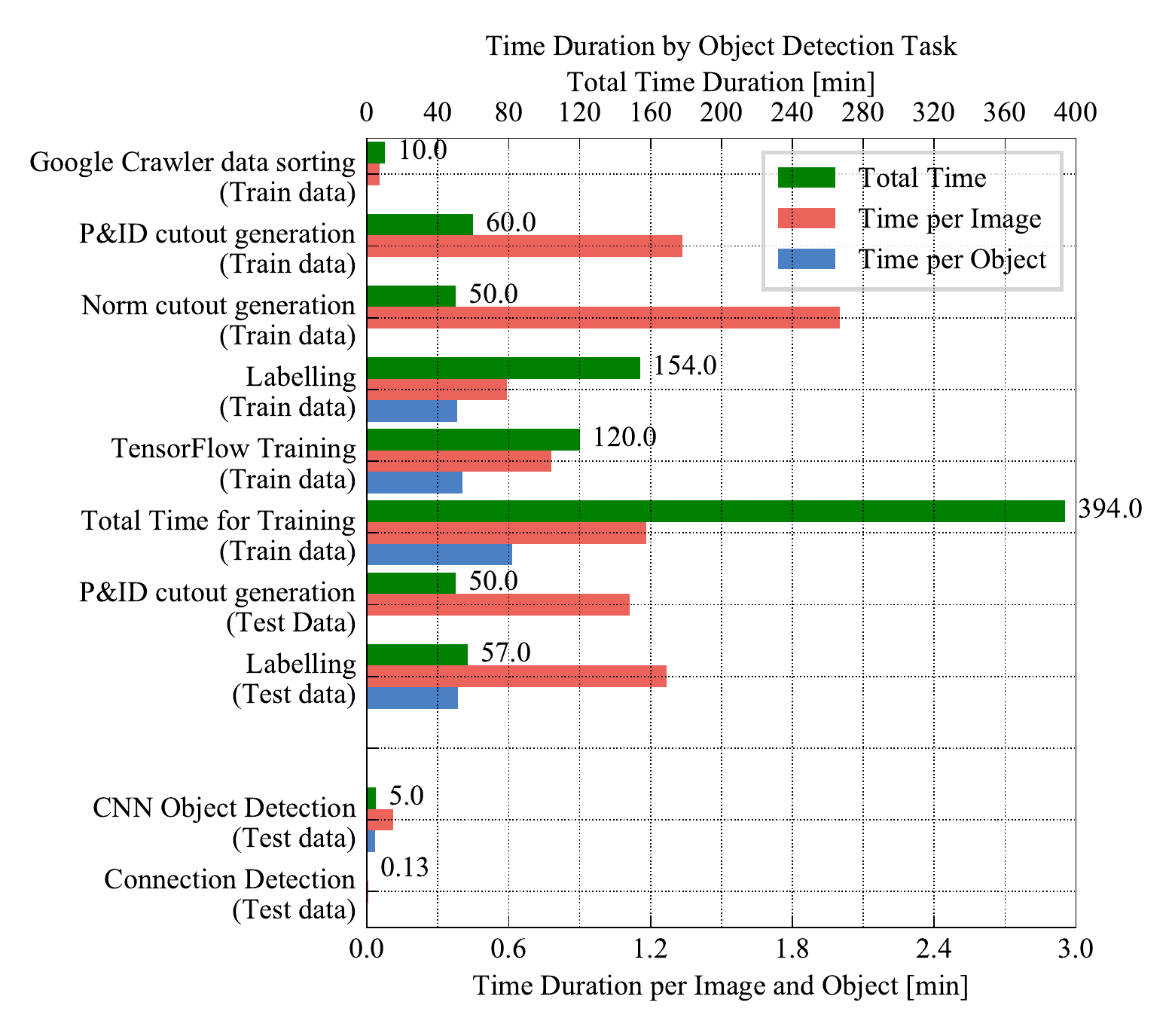}
	\caption{Total time duration, time duration per image and time duration per object for different tasks in object detection}
	\label{fig:TemporalEffortsOD1_2}
\end{figure}

Figure \ref{fig:TemporalEffortsOD1_2} shows the time duration for different procedures of the whole process from data preparation to actual object detection. It shall be noted that these values are not universally valid. They show the time we needed for these tasks and can serve as a reference to evaluate the temporal and effort cost of the different procedures.

Generating the training image set out of the raw data sources is given by the first three points, where only times per image are given. Times per object do not make sense in this case as the work effort is independent of the number of objects drawn on the image. The preprocessing takes us around 120 minutes in total. The time per image ranges from around 0.05 up to 2 minutes. The labeling takes 154 minutes, which is comparable to the preprocessing. The time per object indicates the time needed for labeling one object, which is about 20-25 seconds. The training time is 120 minutes with the data set of 154 images and 296 objects. The times for the described procedures can be summarized to a total time for preparing a usable CNN object detection algorithm, which is around 400 minutes in total. The actual detection process for all images of the test image set takes 5 minutes.

\subsection{Connection detection performance}
\begin{figure}[ht]
	\centering
	\includegraphics[width=0.76\linewidth]{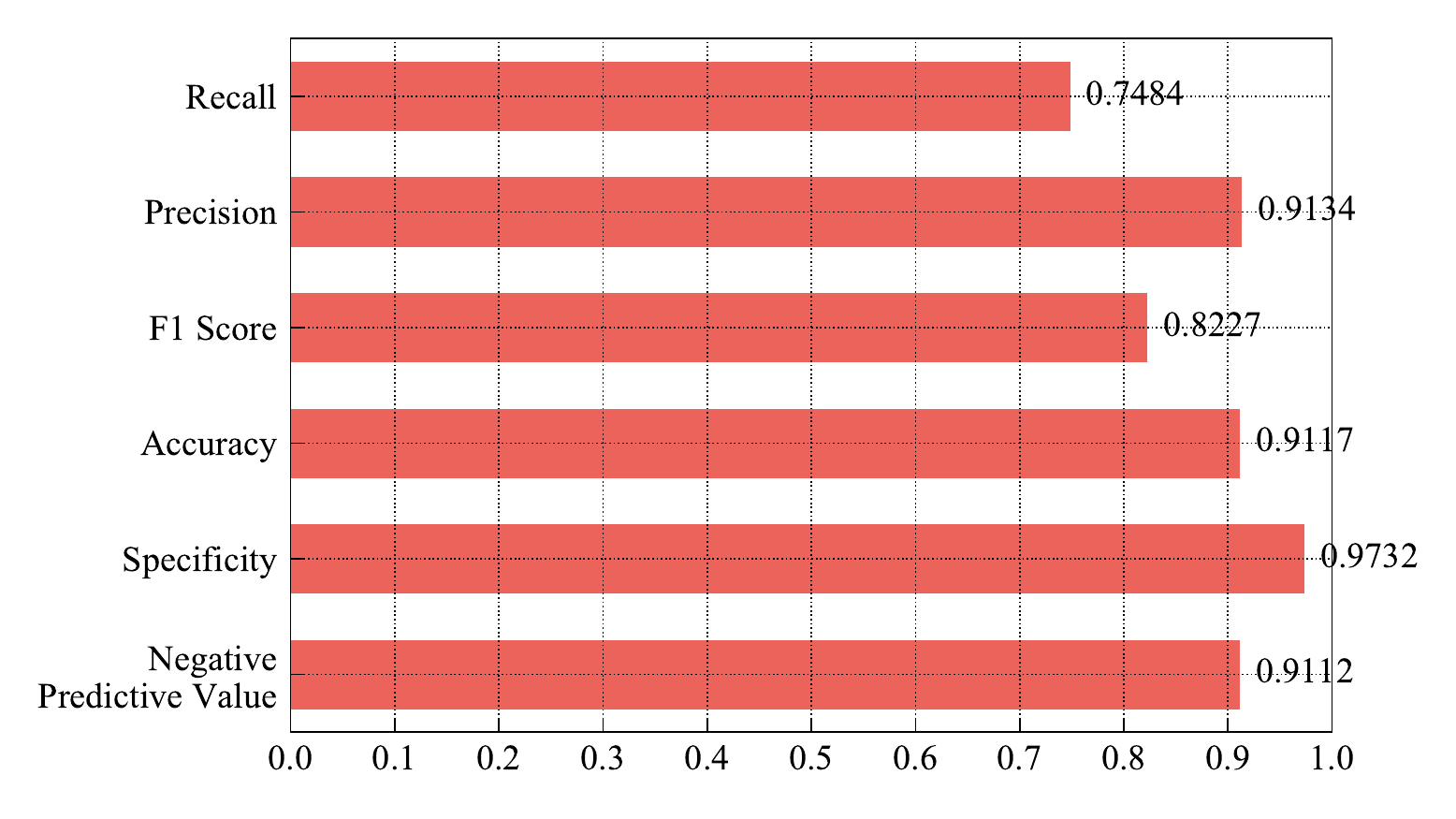}
	\caption{Connection detection evaluation metrics (recall, precision, F1 score, accuracy, specificity, negative predictive value)}
	\label{fig:ConnectionDetEval}
\end{figure}

The connection detection performance is evaluated using the defined connection test data set with the numbers shown in Figure 1 and the ground truth being connection matrices for all test images. To see the performance of the connection algorithm separately without taking into account errors from the object detection, the ground truths for the TBE symbols are used for the evaluation. Figure \ref{fig:ConnectionDetEval} shows the evaluation metrics for the total connection test image data set.
39 elements were identified as false negatives, so were not detected although a connection exists. 116 elements were identified correctly, which leads to a recall of 75\%. The precision is 91\%, 11 elements were misdetected as being connections although no connection exists. The number of false positives is low.

400 connection candidates were predicted correctly as true negative, so correctly not being connections. As the number of true negatives is higher than the number of true positives, the accuracy is higher than the recall, as the accuracy takes both, TP and TN into account. It is slightly above 90\%.

\section{Discussion}

\subsection{Object detection}

Building energy systems (BES), unlike industrial systems, do not have an established standard for the construction of P\&ID diagrams in practice. Despite the inhomogeneous data set, the results of the object recognition are very good with 93\% average precision (AP) over all classes. The results of connection detection are highly dependent on the preceding object detection. However, these results depend on the object class used. Heavily skewed symbols, like the heat exchanger, achieve good results with up to 78\% average precision (AP). However, they do not reach the results of constant proportional symbols like valves with 97.5\% AP. The heat exchanger is a decisive element for the topology detection of building energy systems. Here, two hydraulically decoupled systems are connected with each other. In addition to the skew, the symbols were also marked in different colors and their patterns were different. Other symbols were also different in shape, but could be generalized well by the Faster R-CNN algorithm.

Especially for approaches of other domains, the comparison to our results is difficult. Since there is no open data set for building energy systems, a comparison always depends on the data set used and the object types used. If these conditions are disregarded, the results of object detection are comparable to the results of \cite{YuxiZhang.2019} and \cite{Kang.2019}. \cite{Kang.2019} used template matching. Our results of $<$ 70\% F1 score for template matching show that a universal approach using artificial intelligence is needed. \cite{Nurminen.2020} had problems recognizing small symbols in the P\&ID with Faster R-CNN. We solved this problem by automating the decomposition of the P\&ID in smaller pieces.

Erroneous symbol detections (false positives) were generated especially for symbols that were not included in the considered object types.

\subsection{Connection detection}

With an accuracy of 91\% and a precision of 91\% many connections could be detected correctly. We found out that the algorithm detects two-directional connections very well, but does not achieve this accuracy for connections over three directional crossings and four directional crossings.

Considering all adversities of a comparison, we were able to exceed the results of \cite{Rahul.2019} (65\% accuracy) and performed slightly worse compared to \cite{Kang.2019} (92\% accuracy).

\subsection{Overall evaluation}

It is remarkable that despite the large variety of standards for P\&ID in buildings, our approach produces good results. The actual usefulness in applications depends on the use case.

For the automated transformation of a digital twin in control applications, such as MPC, the recognition of the inputs and outputs of a system is important \cite{Christofides.2013}. The definition of the system boundaries is crucial here. If for MPC the system boundaries are defined to match the plan boundaries, our approach can support MPC. However, if the system boundaries are set differently, a different approach to topology detection must be used. 

For transformed Petri net \cite{Schumacher.2014}, the detected topology can provide a basis for the creation of control code. Here, the correct identification of the connections and the object types is important. The object types could be determined very well. The connections still need manual work for correct use in Petri nets.

For fault detection and diagnosis, the evaluation of our algorithm depends on the methods used. The generation of time series of individual fault models can be supported by our algorithm. In this case, connections that are set individually incorrectly are not decisive. If the automatically generated digital twin is used in a parallel operation for fault detection, the model exported from our approach cannot be used directly. Adjustments must be made here.

The approach in \cite{Stinner.2019c} for generating building energy systems can be used directly. Depending on the use of the model, it is not decisive whether all connections can be correctly identified. For example, in the approach of generic data sets shown in \cite{Stinner.2019c}, the exact connection of the systems is not decisive for the use of the model. However, if the model is used to predict the exact energy consumption (BEPS), an inaccurate model is not suitable \cite{Egan.2018}.

For the integration into a building automation system, e.g. as a simulated virtual sensor, a very good model is necessary, depending on the integration. For this purpose, our approach is only of limited use. Here, the exported digital twin model has to be adapted. However, the translation into a labeling approach for building automation systems supports its use in building automation systems.

\section{Conclusion}

\subsection{Overall performance}
We created a data set for the detection of topologies in building energy systems with piping and instrumentation diagrams as data source. We tracked needed time for steps for preparation of our data set and implementation of the algorithms. The combination of object recognition, line detection, line crossing and connection detection algorithm showed good results. The algorithm recognized all object categories and connections well. An extension of the test data set by further object types could improve the object recognition algorithm. This would result in fewer false detections of object types not taken into account. The integration of skewed symbols in the test data set could also improve the results, especially for the object type heat exchanger.

We showed that plans of technical building equipment are a good source for the system understanding of a building energy system. The direct use of the exported digital twins depends on the application.

\subsection{Further use of the developed digital twins}

The approach presented here provides a base for advanced algorithms to optimize the energy use. We discussed if our approach supports model predictive control, generation of control code, fault detection and diagnosis, automatically generated simulation models and integration into building automation systems depending on the application. We will implement and review these approaches in our tool chain in future work.

\section*{Acknowledgments} 
We gratefully acknowledge the financial support provided by the BMWi (Federal Ministry for Economic Affairs and Energy), promotional reference 03SBE006A.

\bibliography{mybibfile_2}

\end{document}